\pdfoutput=1
\documentclass{article} 
\usepackage{iclr2017_workshop,times}
\usepackage{hyperref}
\usepackage{url}

\usepackage{amssymb,amsmath,amsthm}
\usepackage[makeroom]{cancel}

\usepackage{algorithm}
\usepackage[noend]{algpseudocode}

\usepackage{graphicx}
\usepackage{multirow,bigstrut}
\newcommand{\eqname}[1]{\tag*{#1}}

\usepackage{textcomp}

\usepackage{bm}

\usepackage{subfig} 

\DeclareCaptionSubType*{algorithm}

\definecolor{mydarkblue}{rgb}{0,0.08,0.45}
\definecolor{myfavblue}{rgb}{0.1176, 0.392, 1.0}
\hypersetup{
    colorlinks=true,
    linkcolor=mydarkblue,
    citecolor=mydarkblue,
    filecolor=mydarkblue,
    urlcolor=mydarkblue}

\title{Reinterpreting Importance-Weighted \\Autoencoders}

\author{Chris Cremer, Quaid Morris \& David Duvenaud \\
Department of Computer Science\\
University of Toronto\\
\texttt{\{ccremer,duvenaud\}@cs.toronto.edu} \\
\texttt{\{quaid.morris\}@utoronto.ca}
}

\begin{document}

\maketitle

\begin{abstract}
The standard interpretation of importance-weighted autoencoders is that they maximize a tighter lower bound on the marginal likelihood than the standard evidence lower bound.
We give an alternate interpretation of this procedure: that it optimizes the standard variational lower bound, but using a more complex distribution. 
We formally derive this result, present a tighter lower bound, and visualize the implicit importance-weighted distribution.
\end{abstract}

\begin{figure}[b]
  \centering
   True posterior \qquad  \qquad $k=1$  \qquad \qquad \qquad $k=10$ \qquad \qquad \qquad $k=100$ \quad
      \includegraphics[width=1.\textwidth, clip, trim=2.5cm 3.9cm 2cm 3.6cm]{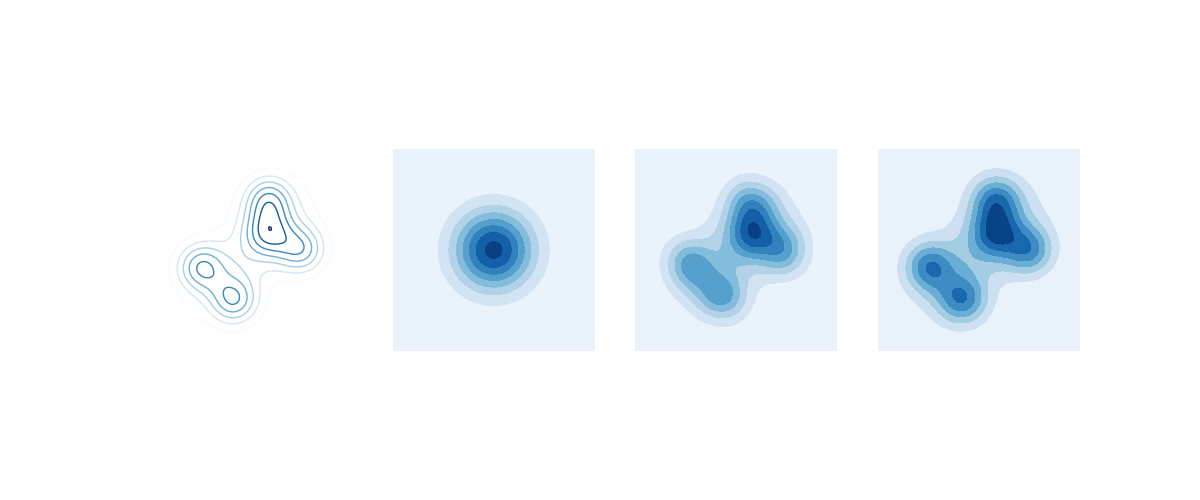}
  \vspace{-5mm}
  \caption{Approximations to a complex true distribution, defined via $q_{EW}$.
  As $k$ grows, this approximation approaches the true distribution.}
  \label{viz1}
\end{figure}

\section{Background}
The importance-weighted autoencoder (IWAE; \cite{burda2015importance}) is a variational inference strategy capable of producing arbitrarily tight evidence lower bounds. IWAE maximizes the following multi-sample evidence lower bound (ELBO): 
\begin{align} 
    \log p(x) &
    \geq E_{z_{1}...z_{k} \sim q(z|x)} \left[\log\left(  \frac{1}{k}  \sum_{i=1}^k \frac{p(x,z_i)}{q(z_i|x)}  \right)  \right] = L_{IWAE}[q] \label{iwae_elbo}  \eqname{(IWAE ELBO)}
\end{align}
which is a tighter lower bound than the ELBO maximized by the variational autoencoder (VAE; \cite{vae}):
\begin{align}
    \log p(x) & \geq E_{z \sim q(z|x)} \left[ \log\left(\frac{p(x,z)}{q(z|x)}  \right)  \right] = L_{VAE}[q]. \label{vae_elbo} \eqname{(VAE ELBO)}
\end{align}




\section{Defining the implicit distribution \texorpdfstring{$\tilde{q}_{IW}$}{}}


In this section, we derive the implicit distribution that arises from importance sampling from a distribution $p$ using $q$ as a proposal distribution. Given a batch of samples $z_{2}...z_{k}$ from $q(z|x)$, the following is the unnormalized importance-weighted distribution:
\begin{align} 
    \tilde{q}_{IW}(z|x,z_{2:k}) =  \frac{ \frac{p(x,z)}{q(z|x)}}{  \frac{1}{k}   \sum_{j=1}^k \frac{p(x,z_j)}{q(z_j|x)}}   q(z|x)
    = \frac{p(x,z)}{\frac{1}{k} \left(  \frac{p(x,z)}{q(z|x)}+ \sum_{j=2}^k \frac{p(x,z_j)}{q(z_j|x)} \right)} 
\label{eq:qiw}
\end{align}

Here are some properties of the approximate IWAE posterior: 
\begin{itemize}
    \item When $k=1$, $\tilde{q}_{IW}(z|x,z_{2:k})$ equals $q(z|x)$.
    \item When $k > 1$, the form of $\tilde{q}_{IW}(z|x,z_{2:k})$ depends on the true posterior $p(z|x)$.
    \item As $k \rightarrow \infty$, $\mathbb{E}_{z_2...z_k} \left[ \tilde{q}_{IW}(z|x,z_{2:k}) \right]$ approaches the true posterior $p(z|x)$ pointwise.
\end{itemize}
See the appendix for details. Importantly, $\tilde{q}_{IW}(z|x,z_{2:k})$ is dependent on the batch of samples $z_{2}...z_{k}$. See Fig. \ref{viz} in the appendix for a visualization of $\tilde{q}_{IW}$ with different batches of $z_{2}...z_{k}$.

\subsection{Recovering the IWAE bound from the VAE bound}

Here we show that the IWAE ELBO is equivalent to the VAE ELBO in expectation, but with a more flexible, unnormalized $\tilde{q}_{IW}$ distribution, implicitly defined by importance reweighting.
If we replace $q(z|x)$ with $\tilde{q}_{IW}(z|x,z_{2:k})$ and take an expectation over $z_2 \dots z_k$, then we recover the IWAE ELBO:
\begin{align}
        \mathbb{E}_{z_{2} \dots z_{k}\sim q(z|x)} \left[ \mathcal{L}_{VAE}[\tilde{q}_{IW}(z|z_{2:k})] \right]
    &= \mathbb{E}_{z_{2} \dots z_{k} \sim q(z|x)} \left[ \int_{z} \tilde{q}_{IW}(z|z_{2:k})  \log\left(\frac{p(x,z)}{\tilde{q}_{IW}(z|x,z_{2:k})} \right)  dz \right] \nonumber \\
    &= \mathbb{E}_{z_{2} \dots z_{k} \sim q(z|x)} \left[ \int_{z} \tilde{q}_{IW}(z|z_{2:k}) \log\left(\frac{1}{k}   \sum_{i=1}^k \frac{p(x,z_i)}{q(z_i|x)} \right)  dz \right] \nonumber \\
    &= \mathbb{E}_{z_{1} \dots z_{k} \sim q(z|x)} \left[  \log\left(\frac{1}{k}   \sum_{i=1}^k \frac{p(x,z_i)}{q(z_i|x)} \right)  \right]  = \mathcal{L}_{IWAE}[q] \nonumber
\end{align}
For a more detailed derivation, see the appendix. Note that we are abusing the VAE lower bound notation because this implies an expectation over an unnormalized distribution. Consequently, we replace the expectation with an equivalent integral.





\subsection{Expected importance weighted distribution \texorpdfstring{ $q_{EW}$}{} }

We can achieve a tighter lower bound than $\mathcal{L}_{IWAE}[q]$ by taking the expectation over $z_2 ... z_k$ of $\tilde{q}_{IW}$. The expected importance-weighted distribution $q_{EW}(z|x)$ is a distribution given by:
\begin{align} 
    q_{EW}(z|x)
    = E_{z_{2}...z_{k} \sim q(z|x)} \left[ \tilde{q}_{IW}(z|x,z_{2:k}) \right] 
    = E_{z_{2}...z_{k} \sim q(z|x)} \left[ \frac{p(x,z)}{  \frac{1}{k} \left( \frac{p(x,z)}{q(z|x)}+ \sum_{j=2}^k \frac{p(x,z_j)}{q(z_j|x)} \right) } \right] \label{marg} 
\end{align}

See section \ref{proof_norm} for a proof that $q_{EW}$ is a normalized distribution. Using $q_{EW}$ in the VAE ELBO, $\mathcal{L}_{VAE}[q_{EW}]$, results in an upper bound of $\mathcal{L}_{IWAE}[q]$. See section \ref{qeiw_proof} for the proof, which is a special case of the proof in \cite{vsmc}. The procedure to sample from $q_{EW}(z|x)$ is shown in Algorithm \ref{sampling_qiw}. It is equivalent to sampling-importance-resampling (SIR).

\subsection{Visualizing the nonparameteric approximate posterior}
The IWAE approximating distribution is nonparametric in the sense that, as the true posterior grows more complex, so does the shape of $\tilde{q}_{IW}$ and $q_{EW}$.
This makes plotting these distributions challenging.
A kernel-density-estimation approach could be used, but requires many samples.
Thankfully, equations \eqref{eq:qiw} and \eqref{marg} give us a simple and fast way to approximately plot $\tilde{q}_{IW}$ and $q_{EW}$ without introducing artifacts due to kernel density smoothing.

Figure \ref{viz1} visualizes $q_{EW}$ on a 2D distribution approximation problem using Algorithm~\ref{plotting_qeiw}.
The base distribution $q$ is a Gaussian.
As we increase the number of samples $k$ and keep the base distribution fixed, we see that the approximation approaches the true distribution. See section \ref{viz_section} for 1D visualizations of $\tilde{q}_{IW}$ and ${q}_{EW}$ with $k=2$.

\begin{figure}[t]
  \centering
      \subfloat
        {
            \begin{minipage}[t]{0.45\columnwidth}
            \begin{algorithm}[H]
            \caption{Sampling $q_{EW}(z|x)$}\label{sampling_qiw}
            \begin{algorithmic}[1]
                \State $\textit{k} \gets \textit{number of importance samples}$
                \For{i in 1...k}
                    \State $z_i \sim q(z|x)$
                    \State $w_i = \frac{p(x,z_i)}{q(z_i|x)}$
                \EndFor
                \State Each $\tilde w_i = w_i/\sum_{i=1}^{k} w_i$
                \State $j \sim Categorical(\bm{\tilde{w}})$
                \State Return $z_j$
            \end{algorithmic}
            \end{algorithm}
            \end{minipage}
        }
        \hspace{-.5cm}
        \qquad \qquad 
      \subfloat
        {
            \begin{minipage}[t]{0.45\columnwidth}
            \begin{algorithm}[H]
            \caption{Plotting $q_{EW}(z|x)$} \label{plotting_qeiw}        
            \begin{algorithmic}[1]
                \State $\textit{k} \gets \textit{number of importance samples}$
                \State $\textit{S} \gets \textit{number of function samples}$
                \State $\textit{L} \gets \textit{locations to plot}$
                \State $\hat f = zeros(|L|)$
                \For {s in 1...S}
                    \State $z_2\dots z_k \sim q(z|x)$
                    \State $\hat{p}(x)=\sum_{i=2}^{k} \frac{p(x,z_i)}{q(z_i|x)}$
                    \For {$z$ in $L$}
                        \State $\hat f[z] \mathrel{{+}{=}} \frac{p(x,z)}{\frac{1}{k} \left(  \frac{p(x,z)}{q(z|x)}+ \hat{p}(x) \right)}$
                    \EndFor
                \EndFor
                \State Return ${\hat f} / S$
            \end{algorithmic}
            \end{algorithm}
            \end{minipage}
        }
\end{figure}



\section{Resampling for prediction}
During training, we sample the $q$ distribution and implicitly weight them with the IWAE ELBO. After training, we need to explicitly reweight samples from $q$.

\begin{figure}[H]
  \centering
    \hspace{-1.6cm} Real \quad  \qquad \qquad \qquad Sample $q(z|x)$  \qquad \qquad \qquad \qquad \qquad  \quad Sample $q_{EW}(z|x)$  \quad \qquad \qquad  
      \includegraphics[width=1.\textwidth, clip, trim=0cm .5cm 0cm 1cm]{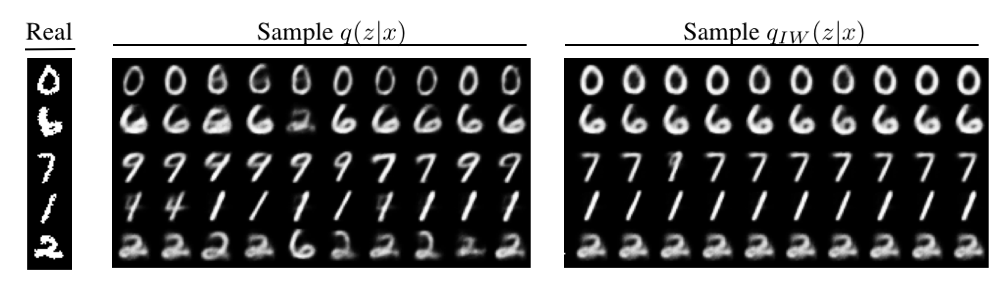}
  \caption{Reconstructions of MNIST samples from $q(z|x)$ and $q_{EW}$.
  The model was trained by maximizing the IWAE ELBO with K=50 and 2 latent dimensions. The reconstructions from $q(z|x)$ are greatly improved with the sampling-resampling step of $q_{EW}$.}
  \label{recon}
\end{figure}

In figure~\ref{recon}, we demonstrate the need to sample from $q_{EW}$ rather than $q(z|x)$ for reconstructing MNIST digits.
We trained the model to maximize the IWAE ELBO with K=50 and 2 latent dimensions, similar to Appendix C in \citet{burda2015importance}.
When we sample from $q(z|x)$ and reconstruct the samples, we see a number of anomalies.
However, if we perform the sampling-resampling step (Alg.~\ref{sampling_qiw}), then the reconstructions are much more accurate.
The intuition here is that we trained the model with $q_{EW}$ with $K=50$ then sampled from $q(z|x)$ ($q_{EW}$ with $K=1$), which are very different distributions, as seen in Fig.~\ref{viz1}.

\section{Discussion}
\cite{bachman} also showed that the IWAE objective is equivalent to stochastic variational inference with a proposal distribution corrected towards the true posterior via normalized importance sampling.
We build on this idea by further examining $\tilde{q}_{IW}$ and by providing visualizations to help better grasp the interpretation.
To summarize our observations, the following is the ordering of lower bounds given specific proposal distributions,
\begin{align} 
    \log p(x) \geq L_{VAE}[q_{EW}] \geq \mathbb{E}_{z_{2} \dots z_{k}\sim q(z|x)} \left[ L_{VAE}[\tilde{q}_{IW}(z|z_{2:k})] \right] = L_{IWAE}[q]  \geq L_{VAE}[q] \nonumber
\end{align}
In light of this, IWAE can be seen as increasing the complexity of the approximate distribution $q$, similar to other methods that increase the complexity of $q$, such as Normalizing Flows \citep{normflow}, Variational Boosting \citep{varboosting} or Hamiltonian variational inference \citep{salimans2015markov}.
With this interpretation in mind, we can possibly generalize $\tilde{q}_{IW}$ to be applicable to other divergence measures.
An interesting avenue of future work is the comparison of IW-based variational families with alpha-divergences or operator variational objectives.

\subsubsection*{Acknowledgments}

We'd like to thank an anonymous ICLR reviewer for providing insightful future directions for this work. We'd like to thank Yuri Burda, Christian Naesseth, and Scott Linderman for bringing our attention to oversights in the paper. We'd also like to thank Christian Naesseth for the derivation in section \ref{qeiw_proof} and for providing many helpful comments.

\bibliographystyle{iclr2017_workshop}
\bibliography{iclr2017_workshop}

\begin{thebibliography}{7}
\providecommand{\natexlab}[1]{#1}
\providecommand{\url}[1]{\texttt{#1}}
\expandafter\ifx\csname urlstyle\endcsname\relax
  \providecommand{\doi}[1]{doi: #1}\else
  \providecommand{\doi}{doi: \begingroup \urlstyle{rm}\Url}\fi

\bibitem[{Bachman} \& {Precup}(2015){Bachman} and {Precup}]{bachman}
Philip {Bachman} and Doina {Precup}.
\newblock {Training Deep Generative Models: Variations on a Theme}.
\newblock \emph{NIPS Approximate Inference Workshop}, 2015.

\bibitem[Burda et~al.(2016)Burda, Grosse, and
  Salakhutdinov]{burda2015importance}
Yuri Burda, Roger Grosse, and Ruslan Salakhutdinov.
\newblock Importance weighted autoencoders.
\newblock \emph{In ICLR}, 2016.

\bibitem[{Jimenez Rezende} \& {Mohamed}(2015){Jimenez Rezende} and
  {Mohamed}]{normflow}
Danilo {Jimenez Rezende} and Shakir {Mohamed}.
\newblock {Variational Inference with Normalizing Flows}.
\newblock \emph{In ICML}, 2015.

\bibitem[{Kingma} \& {Welling}(2014){Kingma} and {Welling}]{vae}
Diederik~P. {Kingma} and Max {Welling}.
\newblock {Auto-Encoding Variational Bayes}.
\newblock \emph{In ICLR}, 2014.

\bibitem[{Miller} et~al.(2016){Miller}, {Foti}, and {Adams}]{varboosting}
Andrew~C. {Miller}, Nicholas {Foti}, and Ryan~P. {Adams}.
\newblock {Variational Boosting: Iteratively Refining Posterior
  Approximations}.
\newblock \emph{Advances in Approximate Bayesian Inference, NIPS Workshop},
  2016.

\bibitem[{Naesseth} et~al.(2017){Naesseth}, {Linderman}, {Ranganath}, and
  {Blei}]{vsmc}
C.~A. {Naesseth}, S.~W. {Linderman}, R.~{Ranganath}, and D.~M. {Blei}.
\newblock {Variational Sequential Monte Carlo}.
\newblock \emph{ArXiv preprint}, 2017.

\bibitem[Salimans et~al.(2015)Salimans, Kingma, and
  Welling]{salimans2015markov}
Tim Salimans, Diederik~P. Kingma, and Max Welling.
\newblock Markov chain monte carlo and variational inference: Bridging the gap.
\newblock \emph{In ICML}, 2015.

\end{thebibliography}

\newpage

\section{Appendix}

\subsection{Detailed derivation of the equivalence of VAE and IWAE bound}
\label{detailed_derivation}

Here we show that the expectation over $z_2 ... z_k$ of the VAE lower bound with the unnomalized importance-weighted distribution $\tilde{q}_{\textnormal{IW}}$, $\mathcal{L}_{\textnormal{VAE}}[\tilde{q}_{\textnormal{IW}}(z|z_{2:k})]$, is equivalent to the IWAE bound with the original $q$ distribution, $\mathcal{L}_{\textnormal{IWAE}}[q]$.

\begin{align}
    \mathbb{E}_{z_{2} \dots z_{k}\sim q(z|x)} \! \left[ \mathcal{L}_{\textnormal{VAE}}[\tilde{q}_{\textnormal{IW}}(z|z_{2:k})] \right] 
    &= \mathbb{E}_{z_{2} \dots z_{k} \sim q(z|x)} \!\left[ \int_{z} \tilde{q}_{\textnormal{IW}}(z|x,z_{2:k})  \log\left(\frac{p(x,z)}{\tilde{q}_{IW}(z|x,z_{2:k})} \right)  dz \right] \\
    &= \mathbb{E}_{z_{2} \dots z_{k} \sim q(z|x)} \!\left[ \int_{z} \tilde{q}_{\textnormal{IW}}(z|x,z_{2:k})  \log\left(\frac{p(x,z)}{\frac{p(x,z)}{\frac{1}{k}   \sum_{i=1}^k \frac{p(x,z_i)}{q(z_i|x)}}} \right)  dz \right] \\
    &= \mathbb{E}_{z_{2} \dots z_{k} \sim q(z|x)} \!\left[ \int_{z} \tilde{q}_{\textnormal{IW}}(z|x,z_{2:k}) \log\left(\frac{1}{k}   \sum_{i=1}^k \frac{p(x,z_i)}{q(z_i|x)} \right)  dz \right] \\
    &= \mathbb{E}_{z_{2} \dots z_{k} \sim q(z|x)} \!\left[\int_z  \tilde{q}_{\textnormal{IW}}(z|x,z_{2:k}) \log\left(\frac{1}{k}   \sum_{i=1}^k \frac{p(x,z_i)}{q(z_i|x)} \right)  dz \right]  \\
    &= \mathbb{E}_{z_{2} \dots z_{k} \sim q(z|x)} \!\left[\int_z  k \frac{\frac{p(x,z)}{q(z|x)}}{\sum_{j=1}^k \frac{p(x,z_j)}{q(z_j|x)}} q(z|x) \log\left(\frac{1}{k}   \sum_{i=1}^k \frac{p(x,z_i)}{q(z_i|x)} \right)  dz \right]  \\
    &= \mathbb{E}_{z_{2} \dots z_{k} \sim q(z|x)} \!\left[\int_{z_1}  k \frac{\frac{p(x,z_1)}{q(z_1|x)}}{\sum_{j=1}^k \frac{p(x,z_j)}{q(z_j|x)}} q(z|x) \log\left(\frac{1}{k}   \sum_{i=1}^k \frac{p(x,z_i)}{q(z_i|x)} \right)  dz \right] \label{change_notation} \\
    &= \mathbb{E}_{z_{1} \dots z_{k} \sim q(z|x)} \!\left[ k  \frac{\frac{p(x,z_1)}{q(z_1|x)}}{\sum_{j=1}^k \frac{p(x,z_j)}{q(z_j|x)}}  \log\left(\frac{1}{k}   \sum_{i=1}^k \frac{p(x,z_i)}{q(z_i|x)} \right)   \right]  \\
    &= \mathbb{E}_{z_{1} \dots z_{k} \sim q(z|x)} \left[  \frac{\sum_{j=1}^k \frac{p(x,z_j)}{q(z_j|x)}}{\sum_{j=1}^k \frac{p(x,z_j)}{q(z_j|x)}}  \log\left(\frac{1}{k}   \sum_{i=1}^k \frac{p(x,z_i)}{q(z_i|x)} \right)  \right] \label{sum_k} \\
    &= \mathbb{E}_{z_{1} \dots z_{k} \sim q(z|x)} \left[  \log\left(\frac{1}{k}   \sum_{i=1}^k \frac{p(x,z_i)}{q(z_i|x)} \right)  \right]  \\
    &= \mathcal{L}_{\textnormal{IWAE}}[q] 
  \end{align}

(\ref{change_notation}): Change of notation $z = z_1$.\\
(\ref{sum_k}): $z_i$ has the same expectation as $z_1$ so we can replace $k$ with the sum of $k$ terms.

\newpage



\subsection{Proof that \texorpdfstring{$q_{EW}$}{} is a normalized distribution}
\label{proof_norm}

\begin{align} 
    \int_z q_{EW}(z|x) dz
    &= \int_z E_{z_{2}...z_{k} \sim q(z|x)} \left[ \tilde{q}_{IW}(z|x,z_{2:k}) \right] dz \\
    &= \int_z E_{z_{2}...z_{k} \sim q(z|x)} \left[ \frac{p(x,z)}{  \frac{1}{k} \left( \frac{p(x,z)}{q(z|x)}+ \sum_{j=2}^k \frac{p(x,z_j)}{q(z_j|x)} \right) } \right] dz\\
    &= \int_z \frac{q(z|x)}{q(z|x)}E_{z_{2}...z_{k} \sim q(z|x)} \left[ \frac{p(x,z)}{  \frac{1}{k} \left( \frac{p(x,z)}{q(z|x)}+ \sum_{j=2}^k \frac{p(x,z_j)}{q(z_j|x)} \right) } \right] dz\\
    &= E_{z \sim q(z|x)} E_{z_{2}...z_{k} \sim q(z|x)} \left[ \frac{\frac{p(x,z)}{q(z|x)}}{  \frac{1}{k} \left( \frac{p(x,z)}{q(z|x)}+ \sum_{j=2}^k \frac{p(x,z_j)}{q(z_j|x)} \right) } \right] \\
    &= E_{z_{1}...z_{k} \sim q(z|x)} \left[ \frac{\frac{p(x,z_1)}{q(z_1|x)}}{  \frac{1}{k} \left( \sum_{j=1}^k \frac{p(x,z_j)}{q(z_j|x)} \right) } \right] \label{nota} \\
    &= k*E_{z_{1}...z_{k} \sim q(z|x)} \left[ \frac{\frac{p(x,z_1)}{q(z_1|x)}}{   \sum_{j=1}^k \frac{p(x,z_j)}{q(z_j|x)}  } \right]\\
    &= \sum_{i=1}^{k} E_{z_{1}...z_{k} \sim q(z|x)} \left[ \frac{\frac{p(x,z_i)}{q(z_i|x)}}{   \sum_{j=1}^k \frac{p(x,z_j)}{q(z_j|x)}  } \right] \label{to_sum} \\
    &= E_{z_{1}...z_{k} \sim q(z|x)} \left[ \frac{\sum_{i=1}^k \frac{p(x,z_i)}{q(z_i|x)}}{   \sum_{j=1}^k \frac{p(x,z_j)}{q(z_j|x)}  } \right] \label{linear} \\
    &= E_{z_{1}...z_{k} \sim q(z|x)} \left[1 \right] \\
    &= 1
\end{align}

(\ref{nota}): Change of notation $z = z_1$. \\
(\ref{to_sum}): $z_i$ has the same expectation as $z_1$ so we can replace $k$ with the sum of $k$ terms. \\
(\ref{linear}): Linearity of expectation.

\newpage

\subsection{Proof that \texorpdfstring{$\mathcal{L}_{VAE}[q_{EW}]$}{} is an upper bound of \texorpdfstring{$\mathcal{L}_{IWAE}[q]$}{}}
\label{qeiw_proof}

Proof provided by Christian Naesseth.


Let $\hat{p}(x|z_{1:k})=\frac{1}{k} \left( \frac{p(x,z)}{q(z|x)}+ \sum_{j=2}^k \frac{p(x,z_j)}{q(z_j|x)} \right)$
\begin{align} 
    \mathcal{L}_{VAE}[q_{EW}] &= E_{z \sim q_{EW}} \left[ log \left( \frac{p(x,z)}{q_{EW}(z|x)} \right) \right] \\
    &= E_{z \sim q_{EW}} \left[ log \left( \frac{p(x,z)}{E_{q(z_{2:k} |x)} \left[ \frac{p(x,z)}{ \hat{p}(x|z_{1:k})} \right]}  \right) \right] \\
    &= E_{z \sim q_{EW}} \left[ log \left( \frac{1}{E_{q(z_{2:k} |x)} \left[ \frac{1}{ \hat{p}(x|z_{1:k}) } \right]}  \right) \right] \\
    &= E_{z \sim q_{EW}} \left[ - log \left( E_{q(z_{2:k} |x)} \left[  \hat{p}(x|z_{1:k})^{-1} \right] \right) \right] \\   
    &= - \int_{z} p(x,z) E_{q(z_{2:k} |x)} \left[  \hat{p}(x|z_{1:k})^{-1} \right] log \left( E_{q(z_{2:k} |x)} \left[  \hat{p}(x|z_{1:k})^{-1} \right] \right) dz \\  
    &\geq - \int_{z} p(x,z) E_{q(z_{2:k} |x)} \left[  \hat{p}(x|z_{1:k})^{-1} log \left(  \hat{p}(x|z_{1:k})^{-1} \right) \right]  dz \label{geq} \\   
    &= - \int_{z} p(x,z) \int_{z_{2:k}} q(z_{2:k} |x) \hat{p}(x|z_{1:k})^{-1} log \left(  \hat{p}(x|z_{1:k})^{-1} \right)  dz \\  
    &= - \int_{z_{1:k}}  \frac{q(z_1|x)}{q(z_1|x)} p(x,z_1)  q(z_{2:k} |x)  \hat{p}(x|z_{1:k})^{-1} log \left(  \hat{p}(x|z_{1:k})^{-1} \right)   dz \label{z1} \\ 
    &= - \int_{z_{1:k}}  \frac{p(x,z_1)}{q(z_1|x)}   q(z_{1:k} |x)  \hat{p}(x|z_{1:k})^{-1} log \left(  \hat{p}(x|z_{1:k})^{-1} \right)   dz \\ 
    &= \int_{z_{1:k}}  \frac{\frac{p(x,z_1)}{q(z_1|x)}}{\hat{p}(x|z_{1:k})}     q(z_{1:k} |x)  log \left(  \hat{p}(x|z_{1:k}) \right)   dz \\ 
    &= k \int_{z_{1:k}}  \frac{\frac{p(x,z_1)}{q(z_1|x)}}{ \sum_{j=1}^k \frac{p(x,z_j)}{q(z_j|x)} }     q(z_{1:k} |x)  log \left(  \frac{1}{k} \sum_{j=1}^k \frac{p(x,z_j)}{q(z_j|x)} \right)   dz \\ 
    &= \sum_{i=1}^{k} \int_{z_{1:k}}  \frac{\frac{p(x,z_i)}{q(z_i|x)}}{ \sum_{j=1}^k \frac{p(x,z_j)}{q(z_j|x)} }     q(z_{1:k} |x)  log \left(  \frac{1}{k} \sum_{j=1}^k \frac{p(x,z_j)}{q(z_j|x)} \right)   dz \label{sum} \\ 
    &=  \int_{z_{1:k}}  \frac{ \sum_{i=1}^{k} \frac{p(x,z_i)}{q(z_i|x)}}{ \sum_{j=1}^k \frac{p(x,z_j)}{q(z_j|x)} }     q(z_{1:k} |x)  log \left(  \frac{1}{k} \sum_{j=1}^k \frac{p(x,z_j)}{q(z_j|x)} \right)   dz \\
    &=  \int_{z_{1:k}} q(z_{1:k} |x)  log \left(  \frac{1}{k} \sum_{j=1}^k \frac{p(x,z_j)}{q(z_j|x)} \right)   dz \\
    &= E_{q(z_{1:k})} \left[ log \left(  \frac{1}{k} \sum_{j=1}^k \frac{p(x,z_j)}{q(z_j|x)} \right)  \right] \\
    &= \mathcal{L}_{IWAE}[q] 
\end{align}

(\ref{geq}): Given that $f(A)=-AlogA$ is concave for $A>0$, and $f(E[x])\geq E[f(x)]$, then $f(E[x])=-E[x]logE[x]\geq E[-xlogx]$.\\ 
(\ref{z1}): Change of notation $z = z_1$.\\
(\ref{sum}): $z_i$ has the same expectation as $z_1$ so we can replace $k$ with the sum of $k$ terms.

\subsection{Proof that \texorpdfstring{$q_{EW}$}{} is closer to the true posterior than \texorpdfstring{$q$}{}}

The previous section showed that $\mathcal{L}_{IWAE}(q) \leq \mathcal{L}_{VAE}(q_{EW})$. That is, the IWAE ELBO with the base $q$ is a lower bound to the VAE ELBO with the importance weighted $q_{EW}$. Due to Jensen\textquotesingle s inequality and as shown in \cite{burda2015importance}, we know that the IWAE ELBO is an upper bound of the VAE ELBO: ${L}_{IWAE}(q) \geq {L}_{VAE}(q)$. Furthermore, the log marginal likelihood can be factorized into: $log(p(x)) = {L}_{VAE}(q) + KL(q||p)$, and rearranged to: $KL(q||p) = log(p(x)) - {L}_{VAE}(q)$.

Following the observations above and substituting $q_{EW}$ for $q$:
\begin{align} 
    KL(q_{EW}||p) &= log(p(x)) - {L}_{VAE}[q_{EW}] \\
    &\leq log(p(x)) - {L}_{IWAE}[q] \\
    &\leq log(p(x)) - {L}_{VAE}[q] = KL(q||p)
\end{align}
Thus, $KL(q_{EW}||p) \leq KL(q||p)$, meaning $q_{EW}$ is closer to the true posterior than $q$ in terms of KL divergence. 

\subsection{In the limit of the number of samples}

Another perspective is in the limit of ${k=\infty}$. Recall that the marginal likelihood can be approximated by importance sampling:
\begin{align} 
    p(x) &= E_{q(z|x)}\left[\frac{p(x,z)}{q(z|x)} \right] \approx \frac{1}{k}\sum_i^k \frac{p(x,z_i)}{q(z_i|x)}
\end{align}
where $z_i$ is sampled from $q(z|x)$. We see that the denominator of $\tilde{q}_{IW}$ is approximating $p(x)$. If $\frac{p(x,z)}{q(z|x)}$ is bounded, then it follows from the strong law of large numbers that, as $k$ approaches infinity, $\tilde{q}_{IW}$ converges to the true posterior $p(z|x)$ almost surely. This interpretation becomes clearer when we factor out the true posterior from $\tilde{q}_{IW}$:
\begin{align} 
\tilde{q}_{IW}(z|x,z_{2:k}) =  \frac{p(x)}{  \frac{1}{k} \left( \frac{p(x,z)}{q(z|x)}+ \sum_{j=2}^k \frac{p(x,z_j)}{q(z_j|x)} \right) }  p(z|x)
\end{align}
We see that the closer the denominator becomes to $p(x)$, the closer $\tilde{q}_{IW}$ is to the true posterior.

\subsection{Visualizing \texorpdfstring{$\tilde{q}_{IW}$}{} and \texorpdfstring{$q_{EW}$}{} in 1D}
\label{viz_section}

\begin{figure}[H]
  \centering
      \includegraphics[width=.5\textwidth]{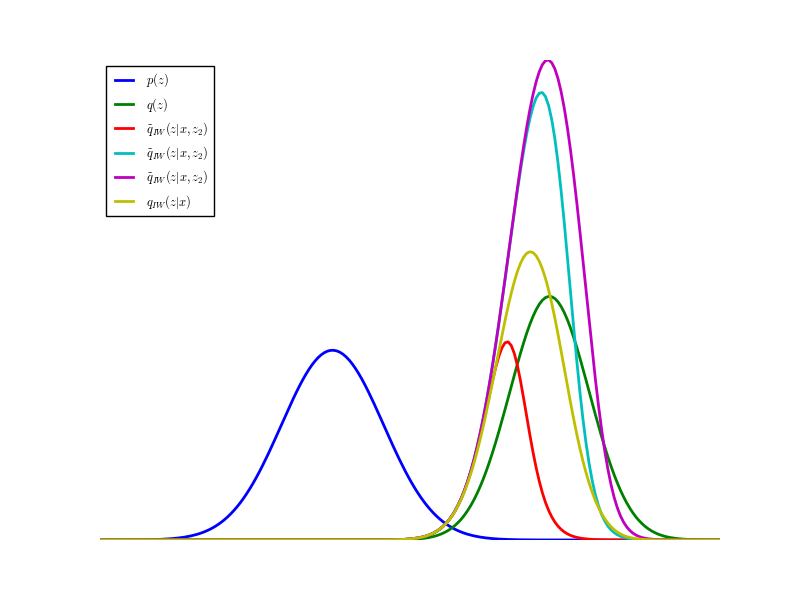}
  \caption{Visualization of 1D $\tilde{q}_{IW}$ and $q_{EW}$ distributions. The blue $p(z)$ distribution and the green $q(z)$ distribution are both normalized. The three instances of $\tilde{q}_{IW}(z|z_2)$ ($k=2$) have different $z_2$ samples from $q(z)$ and we can see that they are unnormalized. $q_{EW}(z)$ is normalized and is the expectation over 30 $\tilde{q}_{IW}(z|z_2)$ distributions. The $\tilde{q}_{IW}$ distributions were plotted using Algorithm \ref{plotting_qeiw} with $S=1$.}
  \label{viz}
\end{figure}

\end{document}